\documentclass[runningheads]{llncs}

\usepackage[mobile]{eccv}

\usepackage{eccvabbrv}

\usepackage{graphicx}
\usepackage{booktabs}

\usepackage[accsupp]{axessibility}  % Improves PDF readability for those with disabilities.

\usepackage[pagebackref,breaklinks,colorlinks,citecolor=eccvblue]{hyperref}
\usepackage{orcidlink}
\usepackage{bbding}

\begin{document}

\title{HO-Gaussian: Hybrid Optimization of 3D Gaussian Splatting for Urban Scenes} 

\titlerunning{HO-Gaussian}

%\author{First Author\inst{1}\orcidlink{0000-1111-2222-3333} \and
%Second Author\inst{2,3}\orcidlink{1111-2222-3333-4444} \and
%Third Author\inst{3}\orcidlink{2222--3333-4444-5555}}
\author{Zhuopeng Li\textsuperscript{1} , Yilin Zhang\textsuperscript{1}, Chenming Wu\textsuperscript{2}, Jianke Zhu\textsuperscript{1*}, Liangjun Zhang\textsuperscript{2}}
\authorrunning{Li et al.}

\institute{Zhejiang University \and Baidu Research}

\maketitle
\begin{figure}[th]
  \centering
  \includegraphics[width=0.8\linewidth]{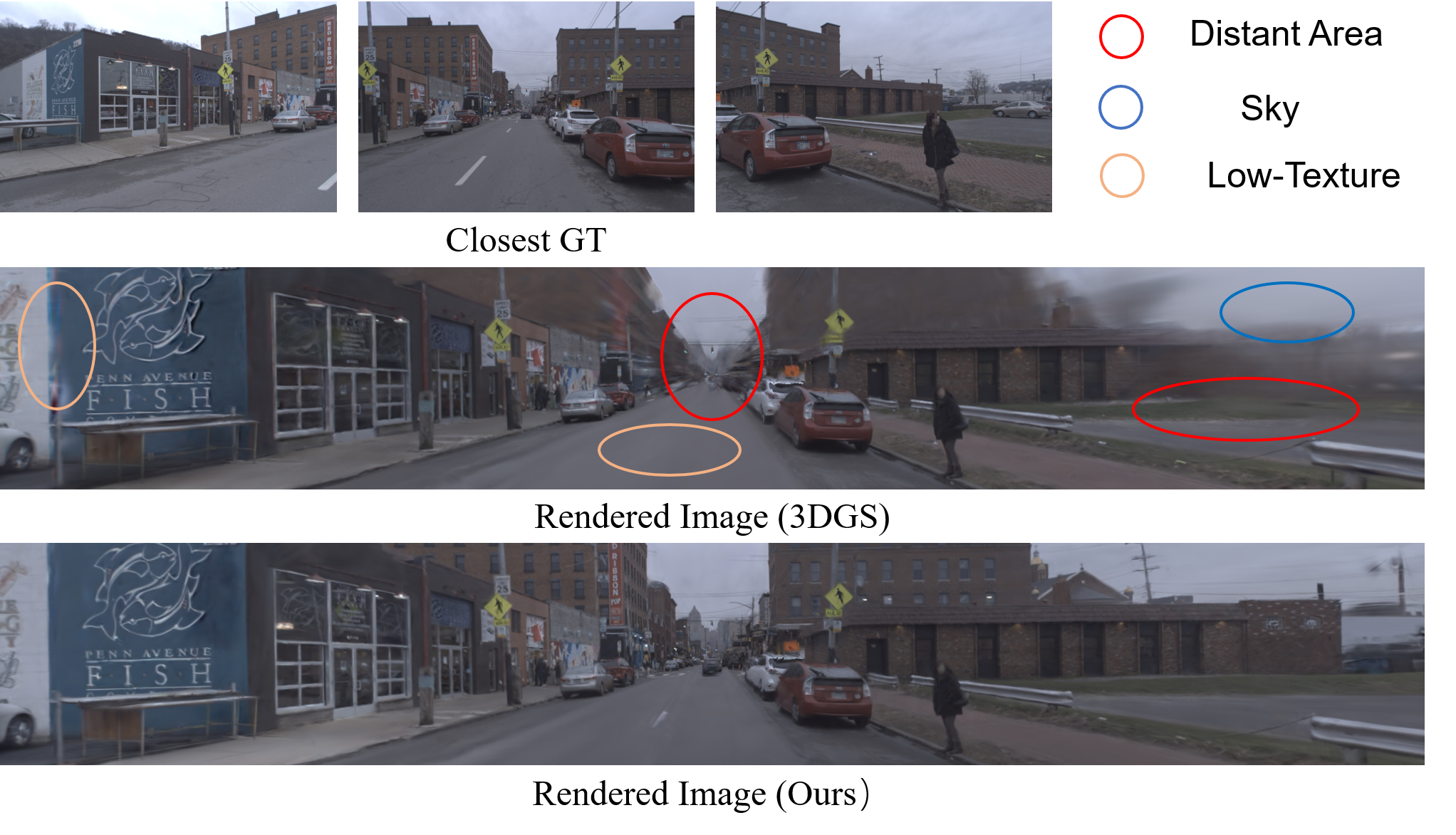}
  \caption{Illustration of 3D Gaussian Splatting(3DGS)~\cite{kerbl20233d} and HO-Gaussian(Ours). Compared with 3DGS initialized by SfM points, our method has richer Gaussian geometric information in low-texture, sky and distant areas, and shows significant improvement in the task of synthesizing novel views.}
  \label{fig:label1}
\end{figure}

\begin{abstract}
The rapid growth of 3D Gaussian Splatting (3DGS) has revolutionized neural rendering, enabling real-time production of high-quality renderings. However, the previous 3DGS-based methods have limitations in urban scenes due to reliance on initial Structure-from-Motion(SfM) points and difficulties in rendering distant, sky and low-texture areas. To overcome these challenges, we propose a hybrid optimization method named \emph{HO-Gaussian}, which combines a grid-based volume with the 3DGS pipeline. HO-Gaussian eliminates the dependency on SfM point initialization, allowing for rendering of urban scenes, and incorporates the Point Densitification to enhance rendering quality in problematic regions during training. Furthermore, we introduce Gaussian Direction Encoding as an alternative for spherical harmonics in the rendering pipeline, which enables view-dependent color representation. To account for multi-camera systems, we introduce neural warping to enhance object consistency across different cameras. Experimental results on widely used autonomous driving datasets demonstrate that HO-Gaussian achieves photo-realistic rendering in real-time on multi-camera urban datasets.

%Experiments on autonomous driving datasets show HO-Gaussian achieves real-time, realistic rendering of multi-camera urban scenes.

\end{abstract}

\section{Introduction}
\label{sec:intro}

Urban scene simulation is of great significance in autonomous driving, which usually generates diverse and large-scale data for training and evaluating various models in autonomous driving, such as occupancy prediction~\cite{zhang2023occnerf,yang2023emernerf}, segmentation~\cite{fan2022nerf,fu2022panoptic,NEURIPS2023_525d2440}, and object detection~\cite{tao2023LiDAR,chang2023neural,li2023adv3d,zhang2023occnerf}.

%In particular, simulating multi-camera scenes can obtain information from multiple perspectives and provide spatially consistent data. \chenming{Not smooth, what do you want to deliver to reviewers here?}

Despite the conventional methods using image-processing techniques to render images in urban scene simulation~\cite{li2019aads, amini2022vista}, recent neural radiance fields (NeRF)-based approaches~\cite{mildenhall2021nerf} have demonstrated remarkable capabilities in facilitating realistic reconstructions and synthesizing novel views from multi-view images. For instance, Block-NeRF~\cite{tancik2022block} decomposes a large scene into blocks and trains NeRFs for each block. DNMP~\cite{lu2023urban} proposes a deformable neural mesh primitives to represent urban scenes by combining mesh-based rendering and neural representations, where each voxel is accurately initialized as a deformable neural mesh primitive. To address the challenge of capturing precise geometry in urban-level scenes, most methods~\cite{ost2022neural,rematas2022urban,turki2023suds,yang2023emernerf} promote the learning of geometry by either incorporating LiDAR observations to supervise NeRF or using LiDAR point clouds as the initialization of the scene~\cite{ost2022neural,xu2022point,you2023nelf}. Nevertheless, synthetic views generated by these methods often exhibit artifacts and lack fine details in texture. Furthermore, the real-time rendering capability of these methods is constrained by the NeRF representation, where volume rendering requires a large number of ray samples per pixel for inference of the MLP network.

In contrast to NeRF, a more recent technique called 3D Gaussian Splatting (3DGS)~\cite{kerbl20233d} offers a notable alternative. 3DGS utilizes explicit 3D Gaussians to represent the scene and has made significant advancements in efficiently rendering novel views. Moreover, 3DGS has been successfully integrated into various platforms and traditional rendering pipelines, including lightweight web rendering and game engines, such as Unity~\cite{haas2014history} and Unreal~\cite{karis2013real}. However, the effectiveness of 3DGS heavily relies on well-initialized point clouds acquired from either SfM (Structure-from-Motion)~\cite{ullman1979interpretation,schonberger2016structure} or SLAM (Simultaneous localization and mapping)~\cite{clemente2007mapping,davison2003real}. While random initialization proves to be effective for object-centric scenes in 3DGS, it becomes inadequate in dealing with unbounded urban environments with limited views. SfM and SLAM systems often treat regions with uniform texture as outliers, resulting in incomplete 3D reconstructions. Additionally, regular LiDAR devices have limited range in providing point cloud data. These factors make it difficult for 3DGS to be optimized effectively, resulting in subpar rendering effects. Moreover, the explicit 3D Gaussian representation occupies large amount of disk space and consumes lots of GPU memory.

%The utilization of SfM and SLAM systems imposes significant computational demands or requires extensive sensor inputs, which can create unnecessary obstacles and hinder the widespread adoption of this technology. 

%To maintain real-time rendering capabilities, we are considering building a rendering pipeline based on Gaussian splatting. We have identified the following challenges: 1) Avoiding the reliance on initializing SfM points in urban scenes using Gaussian-based methods and improving the geometric structure of distant areas, sky, and regions with low-textures; 2) Mitigating the disk space explosion issue associated with Gaussian-based methods in urban scenes; 3) Adapting the Gaussian method for urban scenes with multiple cameras.

To tackle the above challenges in synthesizing urban scenes by 3DGS in real-time, this paper presents HO-Gaussian, a point-free representation method for multi-camera urban scenes. To facilitate end-to-end optimization, we suggest a hybrid scheme to optimize both the volume and Gaussian. Our approach presents point densification to circumvent the reliance on initialization points inherent to 3DGS-based methods. Specifically, we improve rendering quality by introducing a volume to optimize the position of Gaussians, facilitating the learning of geometric information in sky, distant and low-texture areas. To enhance the capabilities of scene representation in the rendering pipeline, we propose a novel Gaussian positional and directional encoding technique that effectively models spherical harmonics and reduces the disk space requirement of Gaussian splating methods. Notably, storing view-dependent spherical harmonic functions even within a range of just a few hundred meters requires gigabytes of disk space in large-scale urban scenes. To render urban scenes with multi-camera, we introduce neural warping scheme that ensures the consistent rendering results across multiple cameras through mutual learning between different camera perspectives. This approach reduces the risk of overfitting to the specific viewpoints, enhancing the generalization capability of the rendering pipeline.
The contributions of this paper can be summarized as follows.

\begin{itemize}
    \item We propose a novel pipeline to learn the positions of Gaussians from a grid-based volume, which allows for better optimization of geometric information and rendering results in urban scenes.
    \item We present the Gaussian positional and directional encoding that improve the scene representation of rendering pipeline, addressing the excessive disk space usage associated with spherical harmonic functions in 3DGS-based methods. Additionally, the presented neural warping scheme enables our approach to efficiently synthesize novel views across various cameras.
    \item Extensive experiments on widely-used autonomous driving datasets demonstrate the effectiveness of our proposed method compared to either the previous NeRF-based methods or 3DGS-based approaches.
\end{itemize}

\section{Related Work}
\subsection{Scene Representation}
3D data can be represented in various forms, such as point clouds, meshes, and voxels. These representations are typically obtained through techniques like Structure from Motion (SfM)~\cite{ullman1979interpretation,schonberger2016structure,cui2017hsfm,smith2016structure,dellaert2000structure}, Multi-View Stereo (MVS)~\cite{schonberger2016pixelwise,kar2017learning,huang2018deepmvs,yao2018mvsnet}, or scanning with LiDAR devices (SLAM systems)~\cite{clemente2007mapping,davison2003real,mur2015orb,mur2017orb,campos2021orb}. SfM and SLAM systems often treat regions with uniform textures as outliers, resulting in incomplete 3D reconstructions. MVS methods, on the other hand, strive to estimate dense 3D geometry while focusing on static scenes and highly overlapping image sets with known poses. However, current neural rendering technologies such as NeRF~\cite{xu2022sinnerf,mihajlovic2022keypointnerf,yang2022ps,tancik2022block,turki2023suds,meuleman2023progressively}, Point-based Rendering~\cite{aliev2020neural,ruckert2022adop,li2023read,sainz2004point,kalaiah2001differential,kopanas2021point,zwicker2001ewa} and 3DGS~\cite{zwicker2001ewa,cheng2024gaussianpro,chen2023text,tang2023dreamgaussian,kerbl20233d} rely on accurate geometric representations to provide clues. To enhance the quality of novel view synthesis in urban scenes, we propose to learn from a grid-based volume and optimize the geometric information through a Gaussian pipeline. This enables us to supplement the geometry information of distant and low-texture areas.

\subsection{NeRF for Urban Scenes}
The implicit neural representation proposed by NeRF has achieved promising results in various scenes, and many researchers have applied it to large-scale outdoor scenes. Block-NeRF~\cite{tancik2022block} first proposed a NeRF composed of multiple regional blocks, using appearance, and exposure embedding to model data in different time periods. SUDS~\cite{turki2023suds} decomposes the scene into three separate hash table data structures to efficiently encode static, dynamic and far-field radiance fields. LocalRF~\cite{meuleman2023progressively} presents an innovative approach by implementing a progressive strategy to dynamically assign local radiance fields. Furthermore, some methods~\cite{yang2023emernerf,lu2023urban,yang2023unisim, wu2023mars, guo2023streetsurf} have achieved promising results by applying NeRF to driving scenarios. However, NeRF-based methods often suffer from artifacts and lack of realistic textures in outdoor scenes. Due to the limited representation capabilities and inefficient rendering pipeline, it is difficult for them to be widely used in the real-world large scale applications.

\subsection{Point-based Rendering and 3D Gaussian Splatting}
Point-based rendering methods are widely used to efficiently render unstructured geometric samples. NPBG~\cite{aliev2020neural} utilizes neural textures to encode local geometric shapes and appearances, enabling high-quality synthesis of novel views from point clouds. Building upon this, ADOP~\cite{ruckert2022adop} proposes a point-based differentiable neural rendering pipeline that leverages single-pixel rasterization to refine all input parameters, such as camera parameters, point cloud positions, and exposure settings.

Recently, 3D Gaussian Splatting~\cite{kerbl20233d} combines the concept of point-based rendering~\cite{aliev2020neural} and splatting~\cite{zwicker2001ewa} techniques for rendering, which employs explicit 3D Gaussian as the representation of the scene. It achieves real-time rendering while maintaining sufficient quality. The explicit representation without a neural network brings fast rendering speed, however, it makes the Gaussian method require a lot of memory and storage resources due to storing Gaussian-related properties, such as covariance matrices and higher-order spherical harmonics. In our work, we employ Gaussian directional encoding as an alternative of spherical harmonics that consume huge disk space.

\begin{figure}[tb]
  \centering
  \includegraphics[width=\linewidth]{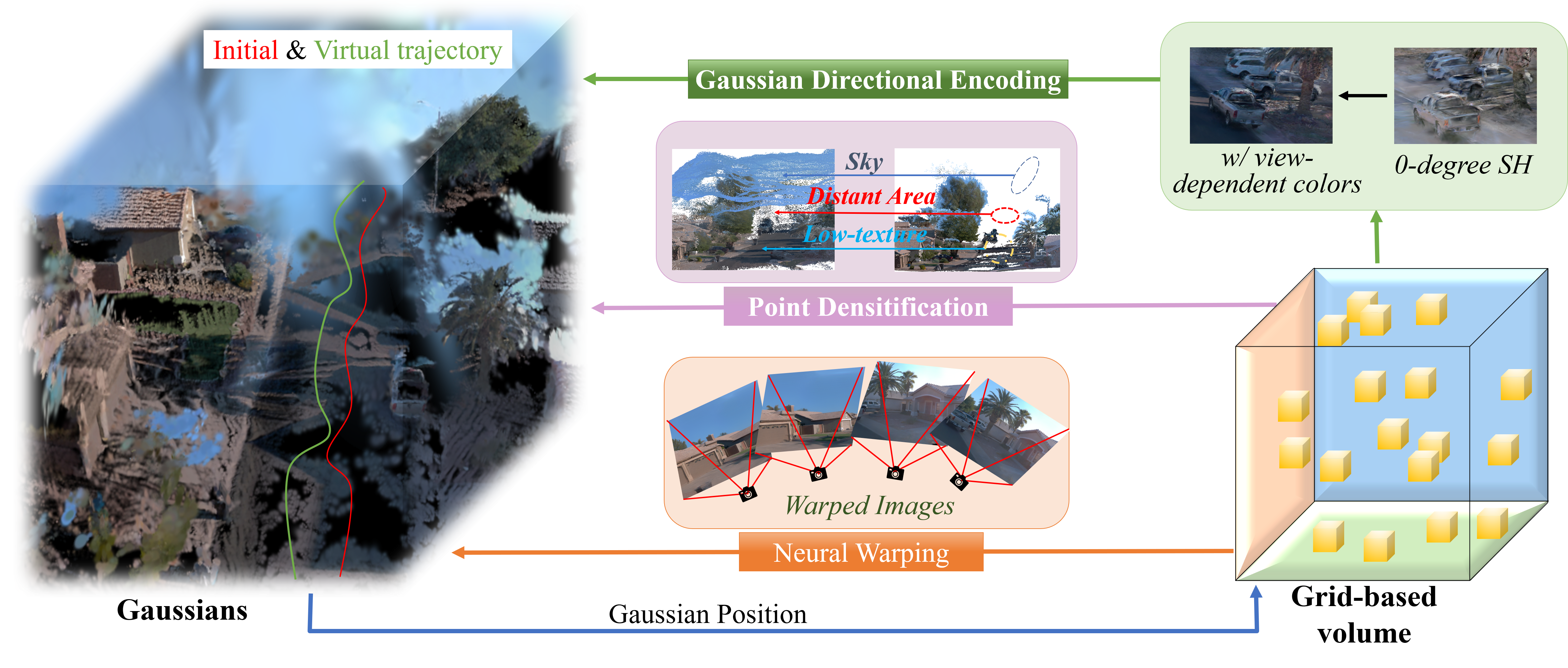}
  \caption{Pipeline. The hybrid optimization starts from a grid-based volume, creating a set of Gaussian points, with the grid-based volume and Gaussian pipeline iteratively optimized. Subsequently, at regular intervals, point densification provides new positions to the Gaussian pipeline to populate problematic regions. Here, view-dependent color is encoded by the Gaussian Directional Encoding, replacing spherical harmonics. Finally, we supply virtual viewpoints to the Gaussian pipeline through neural warping, enhancing consistent appearance and geometry for multi-camera scenes.}
  \label{fig:label2}
\end{figure}

\section{Method}
\label{sec:Method}

Our proposed HO-Gaussian method tackles the challenges of novel view rendering for urban scenes captured by multiple cameras. A key contribution is an end-to-end rendering pipeline grounded in Gaussian splatting, which mitigates the dependency on initial SfM points by employing a grid-based volume. The rendering pipeline facilitates hybrid optimization, bolstering the representation capacity of scene while enriching its geometric information. Crucially, it circumvents the drawbacks of Gaussian splatting in large-scale urban scenarios, such as redundant disk usage, through the design of a grid-based volume representation and Gaussian directional encoding. To enhance the adaptability of rendering pipeline  to multi-camera urban scenes and mitigate the risk of overfitting to specific viewpoints, we introduce a novel neural warping module. Moreover, HO-Gaussian addresses the challenges inherent to multi-camera urban scenes. It achieves real-time rendering performance while preserving photo-realistic texture details.

\subsection{Preliminary}
The 3D Gaussian Splatting method~\cite{kerbl20233d} introduces an explicit representation for 3D scenes that leverages the diverse properties of Gaussians to capture the scene geometry. The approach commences with a set of SfM points and subsequently employs a collection of anisotropic 3D Gaussian models to represent the scene. These Gaussian models inherit various properties inherent to volumetric representations while achieving efficient rendering through a tile-based rasterization algorithm. The 3D Gaussian can be mathematically formulated as:

\begin{equation}G(x)=e^{-\frac{1}{2}(x-\mu)^T\mathbf{\Sigma}^{-1}(x-\mu)},
\label{eq:label1}
\end{equation}
where $x$ denotes an arbitrary position within the 3D scene. Each 3D Gaussian splat is assigned a position (mean) $\mu$, and $\mathbf{\Sigma}$ represents the covariance matrix of the 3D Gaussian. To ensure the positive semi-definiteness of $\mathbf{\Sigma}$, it is represented as the product of a scaling matrix $\mathbf{S}$ and a rotation matrix $\mathbf{R}$:

\begin{equation}\mathbf{\Sigma}=\mathbf{R}\mathbf{S}\mathbf{S}^\mathbf{T}\mathbf{R}^\mathbf{T}.\end{equation}

The Gaussian splatting method leverages splatting techniques ~\cite{zwicker2001ewa} to project the 3D Gaussians onto 2D image planes for rendering purposes. Given the viewing transformation $\mathbf{W}$ and the Jacobian of the affine approximation for the projective transformation $\mathbf{J}$~\cite{zwicker2001surface}, the covariance matrix $\mathbf{\Sigma}^{\prime}$ in camera coordinates can be computed as $\mathbf{\Sigma}^{\prime}=\mathbf{J}\mathbf{W}\mathbf{\Sigma}\mathbf{W}^\mathbf{T}\mathbf{J}^\mathbf{T}.$

Each 3D Gaussian consists of its position, color represented by spherical harmonics (SH), opacity, rotation, and scaling. For a given pixel, it is calculated via multiplying the covariance $\mathbf{\Sigma}$ by the learned opacity $\alpha$ for each point, as shown in Eqn.~\ref{eq:label1}. The color blending of $N$ ordered points with overlapping pixels is determined by:

\begin{equation}\mathbf{C}=\sum_{i\in N}\mathbf{c}_i\alpha_i\prod_{j=1}^{i-1}(1-\alpha_j),\end{equation} 
where $\mathbf{c}_i$ denotes the color of a point, and $\alpha_i$ is its opacity.

\begin{figure}[tb]
  \centering
  \includegraphics[width=\linewidth]{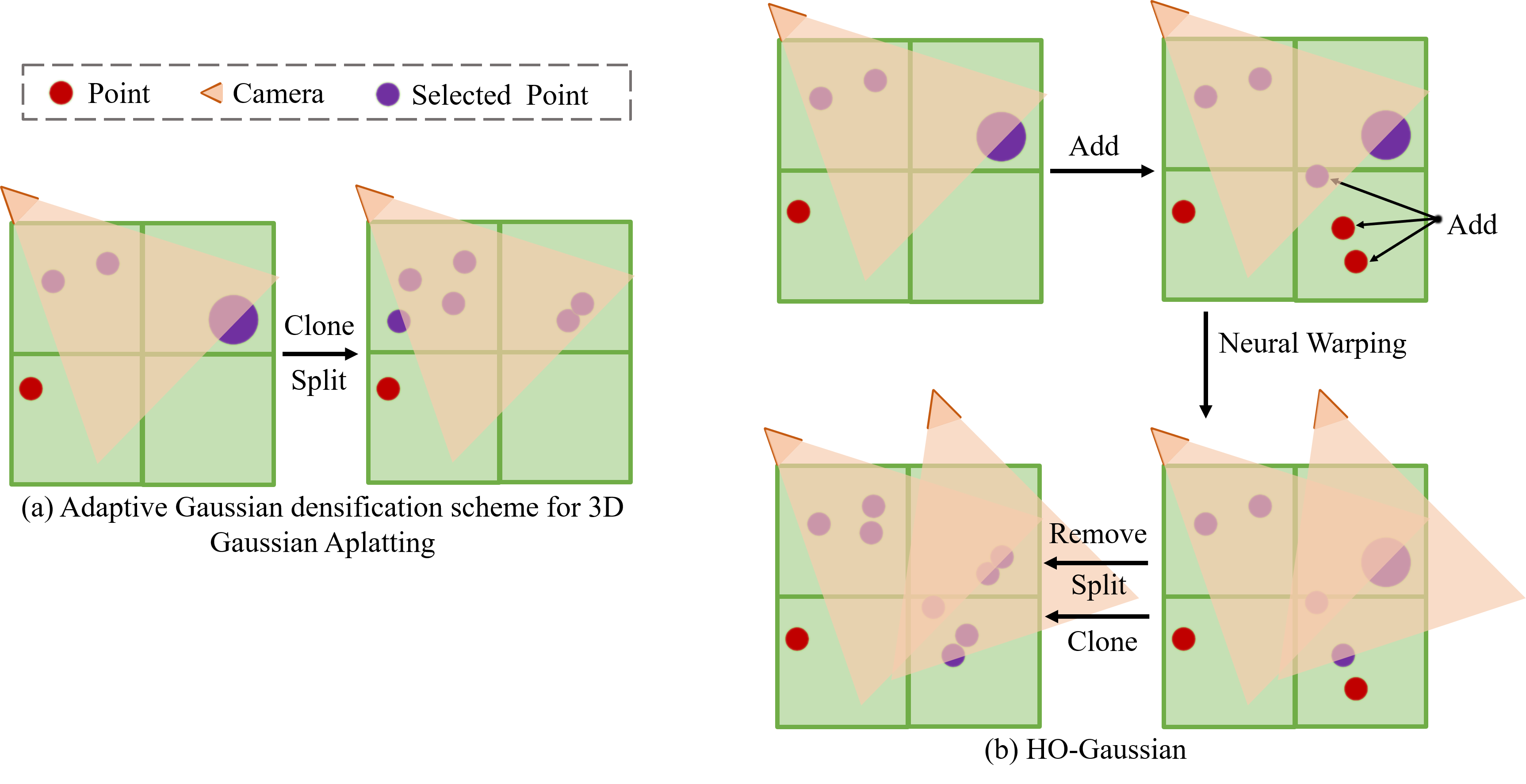}
  \caption{Comparing densification strategies of 3DGS and our HO-Gaussian. The cloning and splitting strategies of 3DGS can effectively optimize the Gaussian distribution near the initial SfM points. However, they fail to work in low-texture or distant areas where the positions of initial points are missing. HO-Gaussian is capable of learning and optimizing Gaussian distributions beyond the initial points. First, Point Densitification supplies the Gaussian pipeline with missing points within viewpoints, preventing the projection of empty 2D splats. Subsequently, Neural Warping introduces virtual viewpoints, thereby covering more occluded points. Finally, clone and split operations are employed to fine-tune the positions of inaccurate splats, and Gaussian splats with opacity values $\alpha$ below a threshold $\epsilon_{\alpha}$ are removed.  }
  \label{fig:label3}
\vspace{-3pt}
\end{figure}

\subsection{Point Densitification based on Hybrid Volume}

The original 3DGS method adaptively controls the number of Gaussian points through clone and split operations. It allows to convert an initial sparse set into a denser one that better represents the scene geometry by filling in empty areas and removing over-reconstructed regions. However, this approach relies on an accurate initialization of the sparse points. Moreover, the original 3DGS densification occurs only in the vicinity of the initial sparse points. For urban scenes, the initial sparse points optimized by SfM techniques often exhibit holes in regions such as the sky, distant areas, or low-texture areas, resulting in blank areas that cannot be rendered accurately. To address this limitation, we introduce a grid-based volumetric representation to provide new positions for the Gaussian pipeline, as illustrated in Fig.~\ref{fig:label3}.

\subsubsection{Point Densitification}

The grid-based volume is a continuous function $f$ mapping location $\mathbf{x}$ and direction $\mathbf{d}$ to a volume density $\sigma\in[0,\infty)$ and color $\mathbf{c}\in[0,1]^{3}$. The sigma  $\sigma_\theta$ and color $\mathbf{c}_\theta$ indicate the density and color prediction of radiance field using MLPs $f_\theta$, parameterized by $\theta$:

\begin{equation}\mathbf{c}_\theta,\sigma_\theta=f_\theta(\mathbf{x},\mathbf{d}).
\label{eq:label5}
\end{equation}

Given a ray $\mathbf{r}$ belonging to the ray set $\mathcal{R}$, we can infer the predicted Gaussian position from grid-based volume by $\mathbf{x}=\mathbf{o}+\hat{\mathbf{D}}_\theta(\mathbf{r})\mathbf{d}_{\mathbf{r}}$, where the 3D position $\mathbf{x}\in\mathbb{R}^3$ and direction $\mathbf{d}\in\mathbb{S}^3$, $\hat{\mathbf{D}}_\theta(\mathbf{r})$ can be approximated by integrating the sampled particles along the ray direction $\mathbf{d}$. $Q(u)$ denotes the accumulated transmittance along the ray%We use the predicted splat position $x$ as the initialization of the Gaussian pipeline.
\begin{equation}\hat{\mathbf{D}}_\theta(\mathbf{r})=\int_{u_n}^{u_f}Q(u)\sigma_\theta(\mathbf{r}(u))udu,\end{equation}
\begin{equation}
Q(u)=\exp\left(-\int_{u_n}^u\sigma_\theta(\mathbf{r}(s))ds\right),\end{equation}
where $u_n$ and $u_f$ is the predefined near and far planes for rendering, respectively. Combined with Eqn.~\ref{eq:label1}, the Gaussian can be expressed as below:

\begin{equation}G(x)=e^{-\frac{1}{2}(\mathbf{o}+\hat{\mathbf{D}}_\theta(\mathbf{r})\mathbf{d}_{\mathbf{r}}-\mu)^T\mathbf{\Sigma}^{-1}(\mathbf{o}+\hat{\mathbf{D}}_\theta(\mathbf{r})\mathbf{d}_{\mathbf{r}}-\mu)}.
\label{eq:label8}
\end{equation}

For stability, we warm up the calculations in the grid-based volume. Subsequently, we select the candidate locations $\mathbf{x}$ within a bounded range and determine whether to add them to the Gaussian pipeline based on whether their density meets a predefined threshold $\tau$. For locations satisfying with the density criterion, we introduce a new Gaussian splat into the Gaussian pipeline, along with its associated Gaussian parameters, such as the corresponding covariance matrix.

%尽管生成的不够准确，通过clone和split来微调

The introduced grid-based volume representation facilitates point densitification, supplying the Gaussian pipeline with missing points within viewpoints, thereby preventing the projection of empty 2D splats, as illustrated in Fig.~\ref{fig:label3}(b). Compared to LiDAR or SfM points, the points generated by this densitification strategy lack precision. To address this limitation, we introduce a hybrid optimization strategy, as described in Section~\ref{sec:Optimization}. Through clone and split operations within the Gaussian pipeline, we fine-tune the positions of inaccurate splats and remove Gaussian splats with opacity values $\alpha$ below a threshold $\epsilon_{\alpha}$.

%Due to the lack of appropriate initial Gaussian parameters, optimization may result in suboptimal local minima, leading to inaccurate estimation of grid-based volume. We integrate the grid-based volume obtained after preliminary training into the Gaussian pipeline. To ensure the spatial continuity of the geometric information provided by the grid-based volume, we compute Gaussian points from the depth estimated by the grid-based volume.

\subsubsection{Gaussian Positional Encoding}
In urban scene, the rich details of large-scale scenes require a significant amount of Gaussian optimization. However, the limitations of video memory for 3D Gaussian result in an uneven distribution of a limited number of Gaussians in space, making it challenging to render the entire scene. Simply applying 3DGS to large-scale scenes can lead to low-quality reconstructions or insufficient memory. Therefore, we propose Gaussian Position Encoding, inspired by Mip-NeRF 360~\cite{barron2022mip}. By compressing the urban scene within a certain range, including nearby objects and those far from the camera, we create a high-quality unbounded scene. By making use of Gaussian positional encoding, we warp the unbounded scene domain into a finite sphere as follows:

\begin{equation}\mathbf{x}_{encoding}=\begin{cases}\mathbf{x}&\operatorname{if}\|\mathbf{x}\|_2\leq1\\\left(2-\frac{1}{\|\mathbf{x}\|_2}\right)\frac{\mathbf{x}}{\|\mathbf{x}\|_2}&\operatorname{if}\|\mathbf{x}\|_2>1\end{cases}\end{equation}

\subsection{The Gaussian Directional Encoding}
Urban scenes encompass a diverse array of elements, including various traffic vehicles, buildings, and other structures, coexisting under different lighting conditions.  The method based on 3DGS requires a large number of Gaussians to model various objects. The rapid increase in the number of Gaussians and the use of high-order SH bring disk Space explodes. Therefore, we utilize view-dependent color ${c}_\theta$ instead of higher-order SH, in which only 0-degree SH is used during training. Specifically, we employ Gaussians to model 3D shapes, while a grid-based neural network generates view-dependent colors as Gaussian directional encodings. Our model adopts the Gaussian representation as: $\{(\mathcal{N}(\mathbf{x}_{encoding}({i}),\mathbf{\Sigma}_{i}),\alpha_{i},c_{i})\}_{i=1}^{n}$, where view-dependent color $c_{i}$ is obtained from the MLPs network $f$ from Eqn.~\ref{eq:label5}.

\subsection{Neural Warping}
Due to the limited overlapping area in multi-camera systems, it is hard to align the colors of objects at different viewpoints, which can make scenes difficult to be optimized. To this end, we introduce a neural warping strategy that simulates images from different virtual viewpoints through grid-based volume, covering various view-dependent colors and virtual positions. This approach helps the model better adapt to real-world scenarios and reduces the risk of overfitting to the specific viewpoints.

By employing a grid-based volume, we generate multiple virtual poses and positions by perturbing the existing pose. This allows us to obtain the warped colors from the volume to assist in fitting objects in multi-cameras through the Gaussian splatting pipeline. Perturbing the pose involves rotating around the current position $x_{p}$ with random angles within the range of $[-10, 10]$. The existing image point captured $p_{i}$ under camera pose $[R_{e},T_{e}]$ is distorted to image point $p_{v}$ with the virtual pose $[R_{v},T_{v}]$. We define the generation of this neural warping as follows
\begin{equation}
p_{v}=K(R_{v}(p_{i})+T_{v}),
\end{equation}
where $K$ is the intrinsic matrix. According to Eqn.~\ref{eq:label5}, we can get the color of the warped image $\mathbf{c_{v}}_\theta=f_\theta(\mathbf{x_{p}},\mathbf{d})$. The warped image provides the Gaussian pipeline with more viewpoints to learn the consistent appearance and geometry of the scene.

\subsection{Hybrid Optimization}
\label{sec:Optimization}
In summary, we optimize the HO-Gaussian pipeline through a hybrid scheme, including Gaussian and volume optimization.

For $N$ Gaussians and their attributes (i.e., position $\mathbf{x}_{encoding}({i})$, opacity $\alpha_{i}$, color ${c}_{\theta{i}}$, covariance matrix $\mathbf{\Sigma}_{i})$), we train the entire model by sampling from ground truth and warped images. We optimize the learnable attribute parameters using a combination of $L_1$ loss and SSIM loss between the ground truth $c(\mathbf{r})$ and rendered images $C$ as below

\begin{equation}\mathcal{L_{\mathrm{g}}}=(1-\lambda)\mathcal{L}_{1}(C,c(\mathbf{r}))+\lambda\mathcal{L}_{\text{SSIM}}(C,c(\mathbf{r})).\end{equation}

The grid-based volume consists of two MLPs with parameters, predicting density and color respectively. Optimize by minimizing the mean square error ground truth $c(\mathbf{r})$ and rendered images $\mathbf{c}_\theta(\mathbf{r})$.

\begin{equation}\mathcal{L}_{\mathrm{MSE}}\left(\theta,\mathcal{R}_\mathbf{r}\right)=\sum_{\mathbf{r}\in\mathcal{R}_\mathbf{r}}\left\|{\mathbf{c}}_\theta(\mathbf{r})-\mathbf{c}(\mathbf{r})\right\|^2.\end{equation}
To jointly optimize all parameters in our proposed Gaussian splatting pipeline, we make use of a combination of multiple losses: 

\begin{equation}\mathcal{L}_\mathrm{total}=\mathcal{L}_\mathrm{g}+\lambda_1\mathcal{L}_{\mathrm{MSE}}\left(\theta,\mathcal{R}_\mathbf{r}\right).\end{equation}

It is worth mentioning that the Gaussian splatting pipeline undergoes the gradient optimization for Gaussian directional encoding, we introduce the supervised learning with hyperparameters $\lambda_1$ to enhance the accuracy of Gaussian points from the grid-based volume.

\begin{figure}[p]
  \centering
  \includegraphics[width=\linewidth]{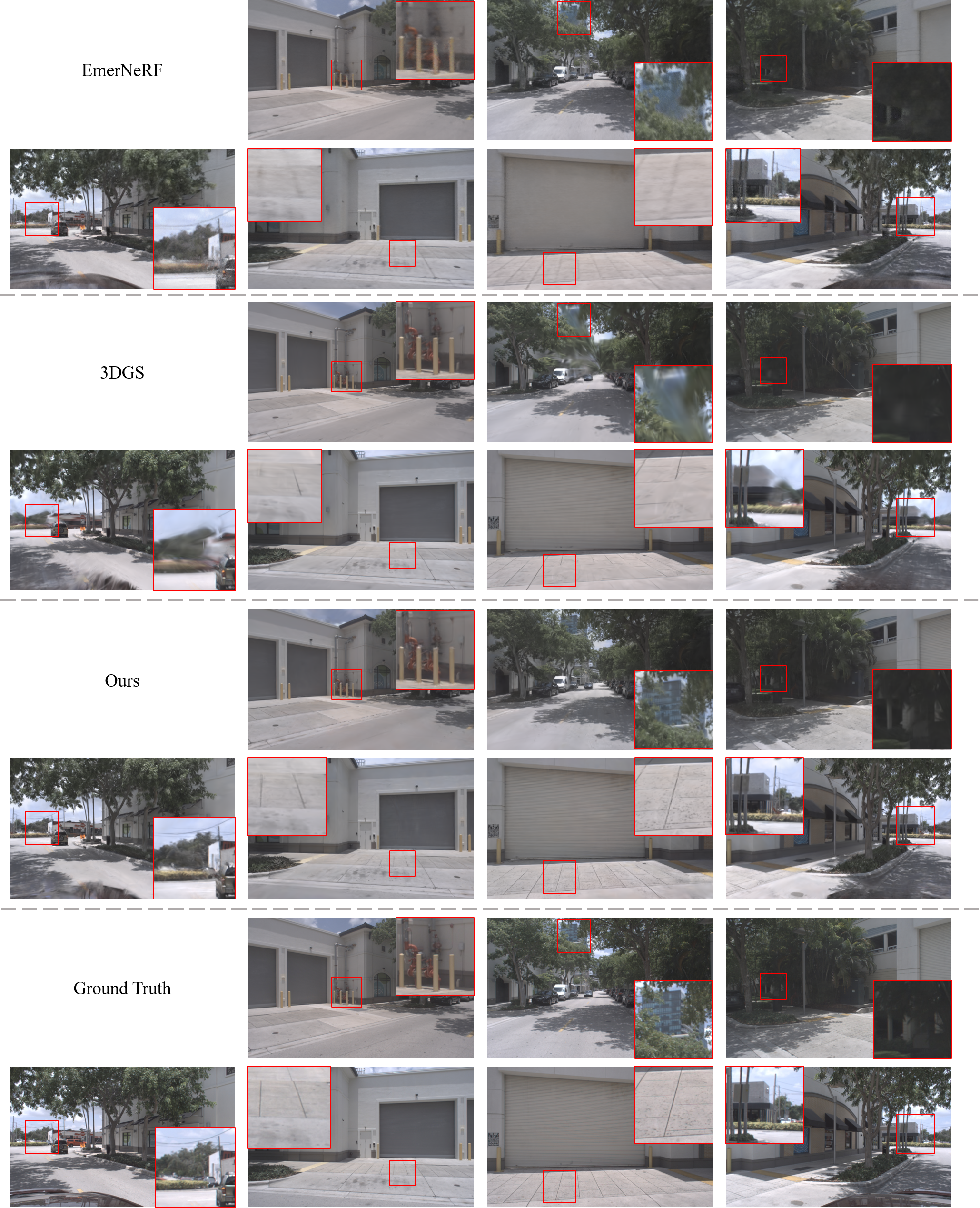}
  \caption{Comparative results of novel view synthesis on Argoverse datasets. Please zoom in to view the detailed results.}
  \label{fig:label5}
\end{figure}

\section{Experimental}
\subsection{Implementation Details}
To evaluate HO-Gaussian, we conducted experiments on an NVIDIA Tesla V100 32GB GPU. HO-Gaussian was rigorously tested on two urban scene datasets of 12 sequences. Through both qualitative and quantitative analysis, we demonstrate convincing results, showing that our method achieves the promising performance and efficiency compared to other methods while mitigating the disk space issues caused by urban scenes. Consistent with the method described in 3DGS, our model retains all hyperparameters of 3DGS and the trained model over 30K iterations in all scenarios. The neural field for view-dependent colors uses a hash grid, followed by an MLP network with a layer number of 2 and 64 channels. The density network consists of an MLP network with a layer number of 1 and 64 channels. We add Gaussian points generated by a grid-based volume to the Gaussian pipeline at 5k, 15k, and 25k training iterations. The values of the hyperparameters $\lambda$ and $\lambda_1$ are set to 0.2 and 0.1, respectively.

\subsection{Datasets}
In our experiments, we evaluated our method on large-scale urban datasets, namely Waymo and Argoverse. We conducted evaluations across eight scenarios from the Waymo dataset and four scenarios from the Argoverse dataset. The Waymo dataset comprises three cameras with a resolution of $1920\times 1280$. We tested our method on both daytime and nighttime scenes from Waymo. As for the Argoverse dataset, it features seven cameras with a resolution of $1920\times 1200$. To thoroughly assess the  representation capability of model and mitigate the potential overfitting issues arising from small scenes, we selectively perform evaluation on scenes containing more than 550 frames of image data. Following~\cite{meuleman2023progressively,tancik2022block,li2023read,aliev2020neural,turki2022mega}, we select every ten frames for testing while choosing the remaining ones for training.

\subsection{Evaluation Results}
 
To demonstrate the effectiveness of the proposed HO-Gaussian method, we compared it with NeRF-based methods for urban scenes that do not require SfM or LiDAR points, including instant NGP~\cite{muller2022instant}, MERF~\cite{reiser2023merf}, Block-NeRF~\cite{tancik2022block} and LocalRF~\cite{meuleman2023progressively}. These methods have shown promising results in synthesizing urban scenes. In order to intuitively reflect the effectiveness of our method, we also compared it against methods that require SfM points or LiDAR information, namely S-NeRF~\cite{xie2023s} and EmerNeRF~\cite{yang2023emernerf}, respectively. Similar to the evaluation protocols used in these methods, we employed three widely used metrics for evaluation: Peak Signal-to-Noise Ratio (PSNR), Structural Similarity Index Measure (SSIM), and Learned Perceptual Image Patch Similarity (LPIPS).

\subsubsection{Comparing Methods without SfM (LiDAR) Point.}

Instant NGP~\cite{muller2022instant} employs hash coding for fast radiance field reconstruction. However, it is unsuitable to reconstruct the large-scale scenes. In this regard, MERF~\cite{reiser2023merf} proposes a novel contraction function to optimize large-scale scenes by distorting the unbounded scene domain into finite space. Significant improvements are observed in both Waymo and Argoverse datasets. Block-NeRF~\cite{tancik2022block} and LocalRF~\cite{meuleman2023progressively} focus on handling outdoor large-scale scene tasks by dividing urban scenes into blocks and employing progressive rendering. Among them, Block-NeRF has poor rendering results due to the poor strategy of dividing multi-camera areas (failed on the Argoverse dataset). LocalRF achieves excellent rendering results in urban scenes, however, it is limited by the scene division strategy in the Waymo dataset. Furthermore, scene division-based NeRF methods often suffer from disk space redundancy, as shown in both Table~\ref{tab:table1} and Table~\ref{tab:table3}.

\begin{table}[tb]
  \caption{Quantitative evaluation of novel view synthesis on the Waymo and Argoverse dataset. "*" denotes half resolution.
  }
  \label{tab:table1}
  \centering
  \begin{tabular}{lccc}
    \toprule
    Method & Input & Waymo  & Argoverse\\
& &  PSNR$\uparrow$ / SSIM$\uparrow$ / LPIPS$\downarrow$ & PSNR$\uparrow$ / SSIM$\uparrow$ / LPIPS$\downarrow$  \\
    \midrule
NGP~\cite{muller2022instant} 
& RGB
& 23.88	/ 0.7369 / 0.5621
& 25.87 / 0.8143 / 0.4364
 \\
MERF~\cite{reiser2023merf} 
& RGB
& 26.28 / 0.7827 / 0.3820
& 27.52 / 0.8473 / 0.3204
  \\
Block-NeRF~\cite{tancik2022block} 
& RGB
&  23.60 / 0.7454 / 0.5031
&  - / - / -
  \\
3DGS~\cite{kerbl20233d} 
& RGB
& 18.31 / 0.6371 / 0.6001
&  21.50 / 0.7979 / 0.4853
  \\
Ours
& RGB
& \textbf{28.03} / \textbf{0.8364} / \textbf{0.3282}
&  30.98 / \textbf{0.9043} / \textbf{0.2287}
\\
Ours*
& RGB
& \textbf{28.97} / \textbf{0.8578} / \textbf{0.2171}
& \textbf{31.52} / \textbf{0.9081} / \textbf{0.1548}
 \\
\midrule
S-NeRF*~\cite{xie2023s} 
& RGB+SfM
& 24.07 / 0.6835 / 0.5215
& 24.36 / 0.7080 / 0.5511 
  \\
EmerNeRF*~\cite{yang2023emernerf} 
& RGB+LiDAR
& 28.62 / 0.8053 / 0.3147
&  30.14 / 0.8347 / 0.3210
  \\  
LocalRF~\cite{meuleman2023progressively} 
& RGB+Depth
& 23.16 / 0.8002 / 0.4201
& \textbf{31.79} / 0.8837 / 0.2976
  \\  
3DGS~\cite{kerbl20233d} 
& RGB+SfM
& 24.90 / 0.8117 / 0.3695
&  27.83 / 0.8795 / 0.2822
  \\
  \bottomrule
  \end{tabular}
    \vspace{-3pt}
\end{table}

\subsubsection{Comparing Methods with SfM(LiDAR) Point.}

In urban-level scenes, most methods rely on using LiDAR or SfM points as scene initialization to reduce unnecessary computations in the radiance field. Alternatively, supervised learning with LiDAR data is used to capture better scene geometry. For example, methods like ~\cite{ost2022neural,rematas2022urban,turki2023suds,yang2023emernerf,lu2023urban}  are commonly employed. In our comparison, we selected two well-performing methods, namely S-NeRF~\cite{xie2023s} and EmerNeRF~\cite{yang2023emernerf}, to assess their performance. Since these two methods require a huge amount of video memory for training, the resolution of the image has to be reduced to half for comparison. Among them, EmerNeRF performs scene decomposition through self-supervision and obtains promising rendering results in various scenes. However, the lack of texture details can be seen from the LPIPS metric, and the rendering speed is limited, as shown in Table~\ref{tab:table1} and Table~\ref{tab:table3}.

In this paper, we advocate for the rendering of large-scale urban scenes without SfM (LiDAR) point initialization or supervision, optimizing geometric information through HO-gaussian and obtaining high-quality rendered images. Among these methods, the Gaussian representation without point cloud initialization performs the worst, as it is challenging for 3DGS-based methods to optimize complex road conditions in large-scale urban scenes without geometric cues. By taking advantage of grid volume learning and Gaussian positions optimization, our proposed approach does not require point cloud initialization and achieves promising rendering results solely based on supervised information from images. As demonstrated in Table~\ref{tab:table1}, our proposed method significantly outperforms both NeRF-based and 3DGS-based methods in almost all evaluation metrics.

\begin{table}[tb]
  \caption{Ablation study.}
  \label{tab:table3}
  \centering
  \begin{tabular}{l c}
    \toprule
     Method  &   PSNR$\uparrow$ / SSIM$\uparrow$ / LPIPS $\downarrow$  \\

    \midrule
           SfM points + 3DGS & 27.83 / 0.8795 / 0.2822 \\
           LiDAR points + 3DGS & 28.40 / 0.8754 / 0.2856  \\ 
           NeRF points + 3DGS &  28.61 / 0.8684 / 0.2883  \\
     \midrule          
           w/ Gaussian directional encoding & 27.70 / 0.8706 / 0.3034  \\
           w/ Gaussian position encoding & 28.42 / 0.8758 / 0.3026 \\
            w/ Point Densitification & 30.58 / 0.8954 / 0.2372   \\         
            w/ Neural warping(Ours) &   30.98 / 0.9043 / 0.2287 \\    
    \midrule
\end{tabular}
  \vspace{-3pt}
\end{table}

\begin{table}[tb]
  \caption{Comparison of Model size, training time and rendering speed.
  }
  \label{tab:table4}
  \centering
  \begin{tabular}{lcccc}
    \toprule
    Method & Input &  Model size &  Training time & FPS \\
    \midrule
S-NeRF*~\cite{xie2023s} 
& RGB+SfM
& 103MB
& 15h
& 0.01
  \\
EmerNeRF*~\cite{yang2023emernerf} 
& RGB+LiDAR
& 431MB
& 57m
& 0.45
  \\  
LocalRF*~\cite{meuleman2023progressively} 
& RGB+Depth
&3762MB
& 14h
& 0.11
  \\  
3DGS*~\cite{kerbl20233d} 
& RGB+SfM
& 557MB
& 31m
& 87
  \\
Ours*
& RGB
& 123MB
& 76m
& 71
\\
  \bottomrule
  \end{tabular}
    \vspace{-3pt}
\end{table}

%waymo, nuscene
% gs with SfM point, with LiDAR point 

\subsection{Ablation study}
%gs, gs+view direction, +nerf point, +warping

%加大场景实验，突出磁盘空间优势
%分开测试，突出端对端加点的优势。有梯度回传，更好的的估计颜色代替sh和生成高斯点

In this section, we conduct qualitative and quantitative experiments using the Argoverse dataset to evaluate each component of our method. We will use Gaussian initialized by SfM points and LiDAR points as the baseline, as shown in Table~\ref{tab:table3}. To demonstrate the advantages of our proposed end-to-end hybrid optimization, we also compare using point clouds generated by NeRF as the initialization of the Gaussian method. It can be seen from the Table~\ref{tab:table3} the Gaussian method initialized by SfM points and LiDAR points performs poorly due to the lack of geometric information in low-texture areas and distance areas, resulting in blurred rendering results, as shown in Fig.~\ref{fig:label1} and Fig.~\ref{fig:label5}. The point cloud generated by NeRF can make up for the shortcomings of low-texture areas. Due to the lack of accuracy, the effect is not much improved.

%enhance rendering quality in problematic regions during training

Compared with the original 3DGS method, we first introduced the view-dependent color based on grid volume to replace the spherical harmonic function, called w/ Gaussian directional encoding. The model size was significantly reduced from 557MB to 123MB, decreasing the disk space usage by 352.8\%. Then, we introduced Gaussian positional encoding, and the results were slightly improved. Through the Point Densitification, we add Gaussian positions to low-texture areas and distant areas of the urban scene, and fine-tune the Gaussian positions by gradient descent of the Gaussian pipeline. As shown in Fig.~\ref{fig:label5} and Table~\ref{tab:table3}, the rendering results of the Gaussian point optimization region have been significantly improved. Finally, we introduce neural warping to provide more virtual viewpoints to the Gaussian pipeline to improve rendering quality.

\begin{figure}[tb]
  \centering
  \includegraphics[width=\linewidth]{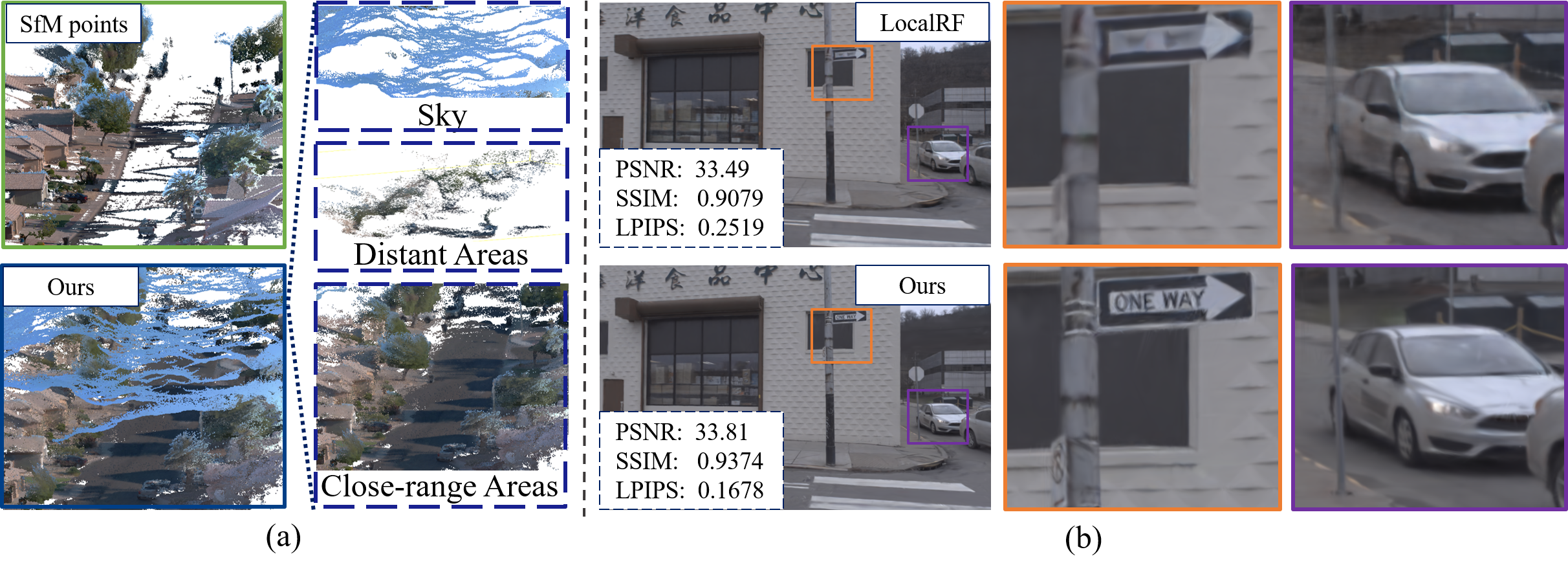}
  \caption{Visualization of scene geometry(a) and texture quality synthesized by LocalRF and our method(b). Please zoom in to view detailed results.}
  \label{fig:label6}
  \vspace{-3pt}
\end{figure}

\subsection{Complexity Analysis}
\subsubsection{Model Size and Rendering Speed}
Rendering speed and model size are crucial for the task of novel view synthesis of urban scenes, since it directly affects interaction efficiency and storage. We evaluate various methods that perform well in large-scale urban scenarios, as shown in Table~\ref{tab:table1} and Table~\ref{tab:table4}. Based on the design of Gaussian directional encoding, our method achieves good rendering quality with smaller disk space and real-time rendering speed.

\subsubsection{Discussion about texture quality and scene geometry}
To demonstrate the effectiveness of geometric optimization in our method, we present geometric predictions in a Waymo scene. Compared with SfM points, the geometry generated by our method has greater advantages in low-texture and distance areas. More results are included in the supplementary material. This section also aims to discuss the reliability of the novel view synthesis evaluation metric. The PSNR score of HO-Gaussian is slightly lower than LocalRF in Table~\ref{tab:table1}. However, the error in the LPIPS metric is reduced by 30.1\%. This difference occurs because PSNR focuses on pixel-level details through mean square error analysis, while the LPIPS metric measures overall image similarity. In Fig.~\ref{fig:label6}, HO-Gaussian exhibits finer texture details compared to LocalRF.

%\subsubsection{Discussion on multi-camera}
%\subsubsection{Discussion about night scene}

%You may also return to \cref{sec:intro} or look at \cref{eq:important}.

\section{Conclusion}
For large-scale urban scenes, this paper proposes a hybrid optimization method that fuses grid-based volume with 3DGS pipeline. HO-Gaussian eliminates the dependency on SfM point initialization and enhances the rendering quality of problematic areas in urban scenes by adding Gaussian points. We introduce Gaussian positional encoding and directional encoding to improve the representation capability of the scene while reducing storage requirement. The HO-Gaussian pipeline is provided with more viewpoints by introducing neural warping to improve the consistent appearance and geometry of the scene. Extensive experiments on several autonomous driving datasets demonstrate its ability to achieve real-time, realistic rendering in multi-camera urban scenes.

\clearpage  % TODO REVIEW/FINAL: This \clearpage needs to be removed from both review and camera-ready versions.

\bibliographystyle{splncs04}
\bibliography{main}
\end{document}